\documentclass[runningheads]{llncs}
\usepackage{graphicx}
\usepackage{amsmath}
\usepackage{multirow}
\usepackage{subcaption}
\usepackage{float}
\usepackage{amssymb}
\usepackage{makecell}
\usepackage{subcaption}
\usepackage{comment}

\begin{document}
\begin{sloppypar}
\title{Progressive Class-level Distillation\thanks{Corresponding email: lijuncst@njnu.edu.cn.}}
\author{Jiayan Li \and
Jun Li\textsuperscript{*} \and
Zhourui Zhang \and Jianhua Xu}
\authorrunning{F. Author et al.}
\institute{School of Computer and Electronic Information, Nanjing Normal University, Nanjing, China, 210023}

\maketitle              % typeset the header of the contribution
\begin{abstract}
%Recently, logit distillation has been receiving increasing attention because of its parameter efficiency and convenience. However, it may lead to suboptimal results. In this paper, we argue that existing logit-based methods may be suboptimal since they all distill logits as a whole. This may overlook those low-probability classes that still contain valuable information. To this end, we proposed a simple but effective method, that is, Difficult-guided Class-level Distillation with Sorted Logits(DCDS). DCDS sorts classes according to their difficulty level and divides the entire logit into multiple subsets of classes. The distillation process is divided into two stages: fine-to-coarse and coarse-to-fine. This helps the student fully utilize the information from the logits, enabling a more comprehensive learning of the teacher's knowledge. Extension experiments on several benchmark datasets demonstrate the effectiveness of DCDS for wide teacher-student pairs. Compared with complex feature-based methods, our DCDS achieves comparable or even better results for image classification and object detection tasks. 
In knowledge distillation (KD), logit distillation (LD) aims to transfer class-level knowledge from a more powerful teacher network to a small student model via accurate teacher-student alignment at the logits level. Since high-confidence object classes usually dominate the distillation process, low-probability classes which also contain discriminating information are downplayed in conventional methods, leading to insufficient knowledge transfer. To address this issue, we propose a simple yet effective LD method termed Progressive Class-level Distillation (PCD). In contrast to existing methods which perform all-class ensemble distillation, our PCD approach performs stage-wise distillation for step-by-step knowledge transfer. More specifically, we perform ranking on teacher-student logits difference for identifying distillation priority from scratch, and subsequently divide the entire LD process into multiple stages. Next, bidirectional stage-wise distillation incorporating fine-to-coarse progressive learning and reverse coarse-to-fine refinement is conducted, allowing comprehensive knowledge transfer via sufficient logits alignment within separate class groups in different distillation stages. Extension experiments on public benchmarking datasets demonstrate the superiority of our method compared to state-of-the-arts for both classification and detection tasks.
%and class-level knowledge is transferred from the teacher to the student with . Subsequent reverse distillation can further enhance fine-to-coarse distillation, which allows bidirectional class-level logits alignment and facilitates comprehensive knowledge transfer. 
%Recently, LD can significantly improve the classification capability of the student model, thereby receiving increasing attention due to its desirable efficiency and adaptability. In particular, the student tends to exhibit deteriorating performance of learning from the teacher when recognizing ``difficult'' classes.  

\end{abstract}
\section{Introduction}
%In recent years, with the remarkable success of deep neural networks (DNNs) in various domains such as image classification~\cite{b1}, speech recognition~\cite{b2}, and natural language processing~\cite{b3}, the scale and complexity of these models have increased substantially. This growth has led to challenges such as high computational cost, memory consumption, and inference latency. Although large models typically deliver superior performance, their slow inference speed, high memory requirements, and intensive resource demands hinder their deployment in real-world scenarios, particularly on resource-constrained devices such as mobile phones, embedded systems, and edge devices. To address these challenges, researchers have proposed a variety of model compression and acceleration techniques, including knowledge distillation~\cite{b4}, pruning~\cite{b5,b6}, quantization~\cite{b7}, and compact model design~\cite{b8,b9,b10}. Among these, knowledge distillation has gained significant attention as an effective method for model compression.

\begin{figure}[t]
    \centering
    \includegraphics[width=\textwidth]{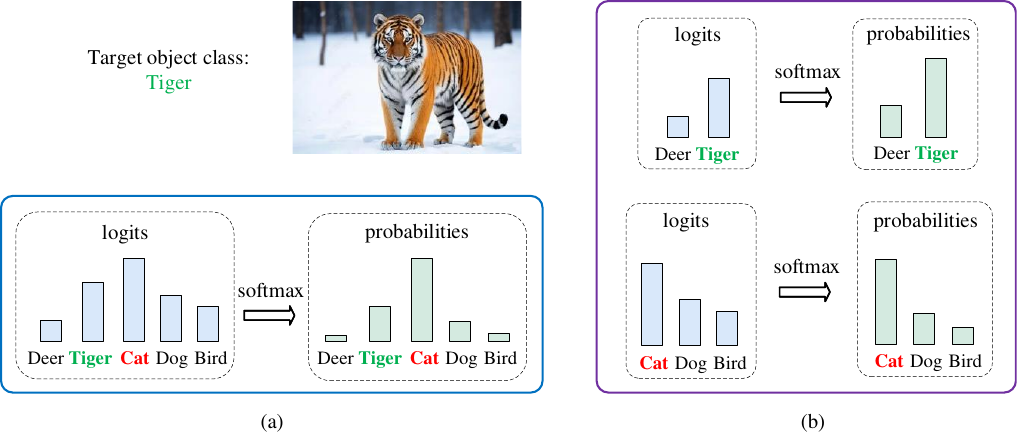}
    \caption{A toy example demonstrating the motivation of our approach. When recognizing a less distinguishable ``tiger'' class, its importance is likely to be understated in conventional distillation methods which largely rely on high-confidence misclassified ``cat'' class, yielding biased knowledge transfer. In contrast, the probability of the target ``tiger'' class is significantly increased within a ``smaller'' class group, and its importance will be elevated in our distillation mechanism.} 
    \label{fig:motivation}
\end{figure}
%it is incorrectly identified as ``cat'', leading to a lower probability score on class ``tiger'' (a). Thus, 
%(a) When an object class is distinctive(e.g. Dog), the predictions tend to be same between teacher and student. (b) When an object class is hard to distinguish(e.g. Tiger), classification results are likely to be divergent. In figure (b), the student incorrectly classified the tiger as a cat, resulting in a very low score for the tiger after softmax function. Consequently, the tiger is likely to be overlooked in the subsequent distillation process.

%While large deep models typically exhibit superior performance nowadays, they usually suffer from inferior efficiency and incur intensive computational demands due to their complex network structures and excessive parameters. This severely hinders their deployment in real-world scenarios, particularly on resource-constrained edge devices. 
As an effective means of model compression, knowledge distillation aims to transfer knowledge from a large and complex teacher network to a smaller and more efficient student model, allowing the latter to achieve competitive performance with compact structure and faster inference speed~\cite{hinton2015distillingknowledgeneuralnetwork}.

In knowledge distillation, mainstream approaches primarily fall into two groups, i.e., feature distillation (FD) and logit distillation (LD). The former enables the student to imitate the teacher features such that the dark knowledge of intermediate network layers can be transferred from the teacher to the student. Different from FD, LD allows the teacher-student alignment at the logit level, such that the classification ability of the student can be promoted. Recently, LD has received extensive attention due to its preferable adaptability and efficiency. It is applicable to different deep frameworks and supplementary to FD for further bridging the gap between the teacher and the student. 
%Earlier LD methods leverage temperature scaling mechanism for reducing the teacher-student difference. Recent research demonstrates that the student performance can be dramatically improved via knowledge decoupling~\cite{b12,b13}. In addition, \cite{b14} and \cite{b15} take advantage of multi-teacher ensemble learning for collaborative distillation. 
In LD, high-confidence object classes predominantly affect the distillation process, while the importance of low-probability classes is understated. The object classes assigned with smaller logits also contain discriminating information that is still crucial for class-specific knowledge transfer. As shown in Fig.~\ref{fig:motivation}, when recognizing an object class that is less distinguishable, misclassified results may be produced with smaller logits on the target class. Thus, its importance is likely to be downplayed in the conventional LD method which essentially performs class-level ensemble distillation, leading to degraded knowledge transfer. 
%Although recent advanced methods demonstrate promising results, they mainly focus on high-confidence object classes without fully exploring the dark knowledge of low-probability classes
%they perform LD for all classes at one time without taking into account significant variances in class-specific recognition difficulty. While recognizing an object class with high distinctiveness, the teacher and the student tend to produce consistent predictions. In contrast, classification results produced from the student network are likely to be deviated from the teacher when identifying an object which is less distinguishable. This class-specific variances lead to the dramatic classification discrepancy between the teacher and the student. As shown in Fig.~\ref{fig:motivation}, variances in the object distinctiveness significantly affect the prediction consistency between the teacher and the student.

%When distilling all-class logits at one time without considering class variance, the sharp concentration property of softmax outputs allows high-probability classes to dominate the distillation process, whilst softmax assigns exponentially suppressed probabilities to classes with relatively smaller logits. 
%While high-probability classes encode abundant object-aware semantics, low-probability classes may still contain discriminative information. In particular, when the student produces biased predictions, correct object class assigned with smaller logits is likely to be downplayed or discarded during knowledge transfer in conventional methods which distill all-class logits at one time. 
To alleviate the above-mentioned drawback, we propose a novel LD method termed Progressive Class-level Distillation (PCD). In contrast to existing ensemble distillation methods, we instead develop a stage-wise distillation strategy for progressive logit-level knowledge transfer. To begin with, we perform ranking on all the object classes in terms of teacher-student prediction discrepancy, leading to a sequence of ``difficult-to-easy'' classes with varying distillation importance. Next, we divide the sequence-guided logit transfer into separate stages such that step-by-step knowledge transfer is achieved. To further refine the transferred knowledge, we perform reverse distillation to achieve bidirectional multi-stage logits alignment. Notably, our proposed PCD method highly resembles the nature of human learning process which incorporates two fundamental stages, i.e., \textbf{knowledge accumulation for progressive learning and knowledge verification for adaptive refinement}. The former enables progressive learning process for step-by-step knowledge transfer, while the latter allows reverse learning for refining knowledge acquisition with improved adaptability. To summarize, main contributions of this study are threefold as follows:

\begin{itemize}
    \item We reveal that low-probability classes containing discriminating information are downplayed in the conventional distillation methods, and should be fully explored for achieving comprehensive knowledge transfer.
    %different object classes exhibit varying discriminating characteristics, leading to significant discrepancy in the teacher-student classification results. The class-specific variances pose great challenges to logits-level knowledge transfer, particularly when the student produce biased predictions. %Conventional LD methods perform all-class ensemble distillation without taking into account this class-specific variances, which is detrimental to logits-level knowledge transfer. treating the logits as a whole during distillation leads to information loss, which hinders the student model from learning a more comprehensive representation of knowledge
    \item We develop a simple yet effective LD method termed Progressive Class-level Distillation (PCD). In contrast to existing methods which perform all-class ensemble distillation, our PCD achieves bidirectional stage-wise distillation which incorporates both fine-to-coarse learning and reverse coarse-to-fine learning for step-by-step knowledge transfer. %DCDS follows two key processes: knowledge accumulation and knowledge verification, which allow the student models to fully exploit the information contained in the logits and to examine and refine the acquired knowledge at multiple levels of granularity. This approach enhances the student’s understanding and generalization capabilities.
    \item Extensive experiments for classification and detection tasks demonstrate the superiority of our method over current mainstream LD techniques, suggesting the promise of our proposed progressive distillation mechanism. %on two standard classification datasets and one object detection dataset show that our method achieves superior performance over the mainstream distillation techniques under various model setups, confirming its effectiveness and versatility.
\end{itemize}

The remainder of this paper is structured as follows. After briefly reviewing the related work in Section~\ref{sec2}, we will elaborate on our proposed PCD method in Section~\ref{sec3}. Extensive experimental evaluations will be carried out in Section~\ref{sec4} before the paper is concluded in Section~\ref{sec5}.

\section{Related Work}\label{sec2}
In this section, we will briefly review knowledge distillation which can be categorized into the following two groups, i.e., FD and LD.
%Knowledge distillation was first proposed by Hinton et al.~\cite{b4}, aiming to train a lightweight student model under the supervision of a large teacher model. The teacher generates soft targets from the input image, which are then used to guide the learning process of the student model. By fitting the logits produced by the teacher, the student can capture richer inter-class relational information, thereby improving its performance. With continued research, a growing number of knowledge distillation methods have emerged to further enhance the learning ability of student models. Currently, existing knowledge distillation methods can be broadly categorized into two types.

\subsection{Feature Distillation}
Feature distillation essentially achieves feature imitation that is capable of transferring feature knowledge of the teacher model to the student model. As the earliest FD model, FitNet~\cite{romero2015fitnetshintsdeepnets} reveals that intermediate layers of the teacher network contain abundant local patterns and structural information, which can also be learned by the student network. Following FitNet, numerous methods have been proposed to take advantage of intermediate features for improving representational capacity of the student with the help of local-global decoupling~\cite{yang2022focal}, feature masking~\cite{yang2023amd,zhang2023structured}, scale-aware adaptive learning~\cite{zhang2025samkdspatialawareadaptivemasking} and contrastive learning~\cite{tian2019contrastive}.

\subsection{Logit Distillation}
Different from FD, LD primarily focuses on logit-level knowledge transfer such that the student model can mimic output probability distribution of the teacher model. Benefiting from temperature scaling mechanism, earlier LD scheme can effectively bridge the teacher-student difference. Recent years have witnessed significant progress in LD research. Jin et al.~\cite{jin2023multi} introduced a multi-level logit distillation method, which allows comprehensive knowledge transfer at different hierarchical levels. Zhao et al.~\cite{zhao2022decoupled} decomposed traditional distillation loss into target and non-target class components each of which is handled with appropriate importance. Wei et al.~\cite{wei2024scaled} decoupled the global logit output into multiple local outputs and construct separate distillation pipelines, achieving fine-grained and unambiguous knowledge transfer. Sun et al.~\cite{sun2024knowledge} utilized masking strategy to suppress incorrect knowledge transfer. Li et al.~\cite{li2023curriculum} applied curriculum learning principles to progressively increase the difficulty of distillation by dynamically adjusting the temperature parameter. Compared to FD, LD methods are more computationally efficient, and easier to be integrated into other distillation frameworks.

\section{Methods}\label{sec3}
\subsection{Background}
Before elaborating on our proposed method, notations regarding conventional LD framework are presented as follows. Mathematically, the output logits of a single sample are denoted as $\mathbf{z} \in \mathbb{R}^{C}$, where \(C\) indicates the total number of object categories in dataset. By applying the softmax function with a temperature factor $\tau$, the softened classification probability distributions $p$ and $q$ for the teacher and student are computed as:
\begin{equation}
p_i = \frac{\exp(z_i^t / \tau)}{\sum_{j=1}^{C} \exp(z_j^t / \tau)}, \quad
q_i = \frac{\exp(z_i^s / \tau)}{\sum_{j=1}^{C} \exp(z_j^s / \tau)}     
\end{equation}
where $z_i^{t/s}$ indicates the logit output of $i^{th}$ class predicted by the teacher/student model, while $\tau$ denotes the temperature that controls the smoothness of the probability distribution.

Consisting of a standard cross-entropy (CE) loss and a Kullback-Leibler (KL) divergence loss, vanilla LD loss function is formulated as:
\begin{equation}
    \mathcal{L}_{KD} = \alpha \cdot \mathcal{L}_{CE} + \beta \cdot \mathcal{L}_{KL} = \alpha \cdot \text{CE}(\mathbf{q}_{\tau=1}, \mathbf{y}) + \beta \cdot \tau^2 \cdot \text{KL}(\mathbf{p}_\tau, \mathbf{q}_\tau)
\end{equation}
where $\alpha$ and $\beta$ are hyper-parameters that balance the two terms. 
\begin{figure}
    \centering
    \includegraphics[width=\textwidth]{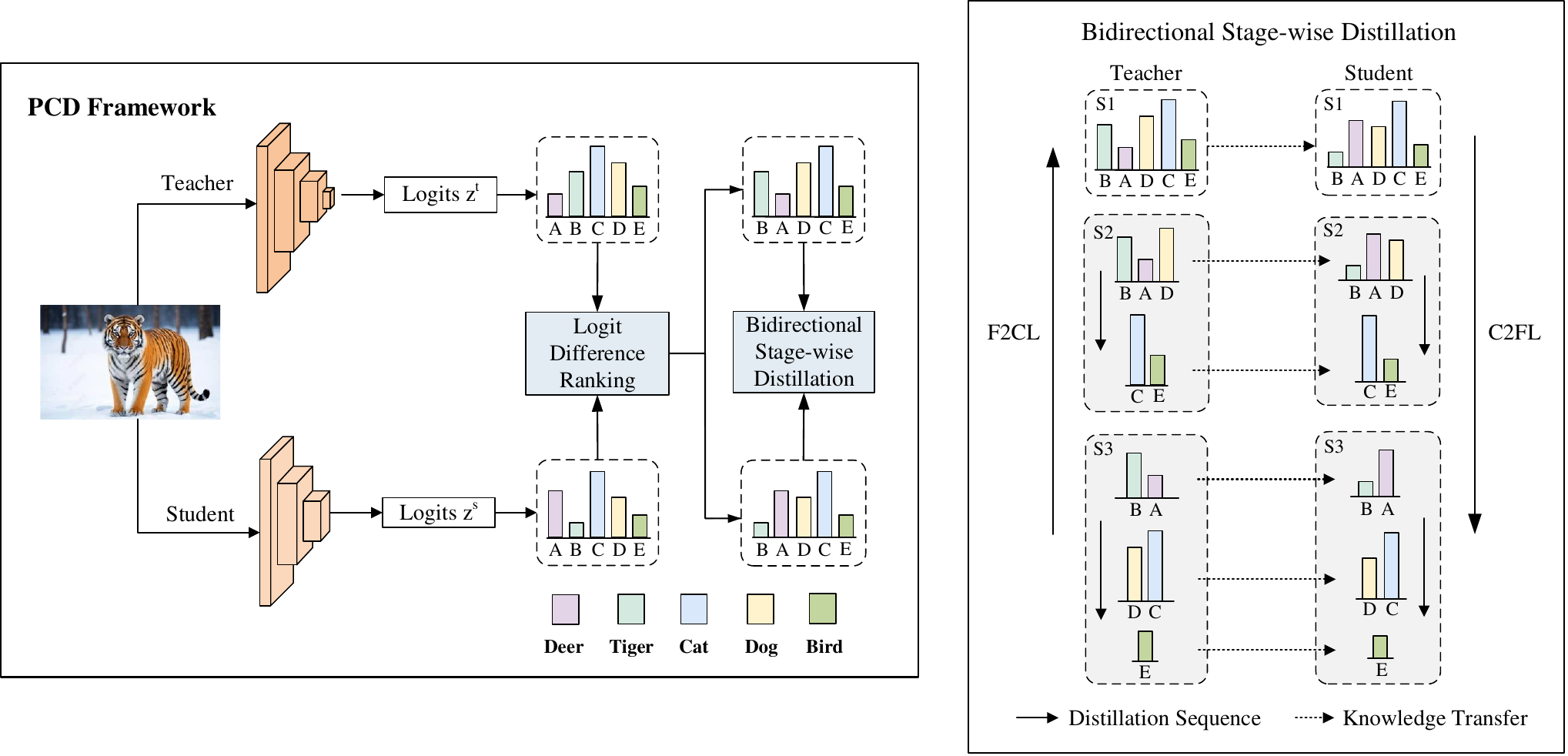}
    \caption{Illustration of our PCD which comprises two components including Logit Difference Ranking (LDR) and Bidirectional Stage-wise Distillation (BSD). LDR performs ranking on the logit difference between the teacher and student, leading to a class sequence indicating distillation priority. Based on the sequence, BSD achieves bidirectional logit alignment with Fine-to-Coarse progressive Learning (F2CL) and reverse Coarse-to-Fine Learning processes (C2FL), achieving comprehensive knowledge transfer.}
    \label{fig:PCD}
\end{figure}
%Compared with the traditional KD that treat the logits as a whole for one-shot distillation, DCDS can make distillation more refined and fully utilize the information contained in logits by difficult-guided sorting and class-level distillation. In the figure, we provide an example with the \(level\) set to 3 to demonstrate this process.

\subsection{Progressive stage-wise distillation}
In this section, we will introduce the proposed PCD framework. Taking into account teacher-student prediction variances, our approach fully explores low-confidence object classes via progressive LD for comprehensive knowledge transfer. As shown in Fig.~\ref{fig:PCD}, PCD consists of two critical modules including Logit Difference Ranking (LDR) and Bidirectional Stage-wise Distillation (BSD) both of which will be discussed in detail as follows.

\subsection{Logit Difference Ranking}
As aforementioned, biased classification results are likely to be produced when recognizing a less distinguishable object. In addition, significant discrepancies often exist between the teacher-student predictions in varying object classes. In particular, the student model is likely to produce biased predictions deviated from the teacher when identifying difficult classes, which implies that it is preferable to handle different classes with varying importance in the distillation process. To emphasize the ``hard-to-learn'' classes, we propose to perform ranking on teacher-student logit difference in a descending order and the classes with notable variances in teacher-student prediction results could be prioritized in logits transfer. This is similar to the process of knowledge acquisition where difficult knowledge is key learning point and merits special attention. 
%In the LD process, significant discrepancies often exist between the predictions of the teacher and the student in different classes. To more effectively guide the student model to focus on hard-to-learn or error-prone categories, we propose the Sorting Module (SM). This mechanism guides the student to focus on the difficult categories and gradually distill to the easier ones, thereby accomplishing efficient knowledge accumulation.

Mathematically, given a batch of samples with teacher and student logit outputs respectively denoted as $\mathbf{z}^t \in \mathbb{R}^C$ and $\mathbf{z}^s \in \mathbb{R}^C$, their class-wise prediction differences are calculated and sorted in a descending order as follows:
%\begin{equation}
%\text{d} = |z^t - z^s|
%\end{equation}

\begin{equation}
    I = \text{argsort} \left( | \mathbf{z}^t-\mathbf{z}^s| , \; \downarrow \right)
\end{equation}
where $I$ denotes the sequence of sorted class index and $C$ indicates the total number of object classes. After obtaining the sequence which encodes varying degrees of teacher-student differences, more challenging classes corresponding to top-ranked indices are prioritized in distillation process. 

\begin{comment}
   After sorting, we can select the classes to be distilled at each step based on the sorted indices:
    \begin{equation}
    M_{i}[b,c] =
    \begin{cases}
        1& \text{if } \text{start}_i \leq \text{I}[b,c] < \text{end}_i, \\
        0& \text{otherwise}
    \end{cases}
    \label{Eq5}
    \end{equation}
    where \( M_{i}[b,c]\) denotes the index mask, which is used to indicate the subset of classes involved in each distillation step. 
\end{comment}

\subsection{Bidirectional Stage-wise Distillation}
%As shown in Fig.~\ref{fig:PCD}, with sorted logits, we propose the class-level disitllation to align teacher output and student output from two processes: fine-to-coarse and coarse-to-fine.
To achieve progressive LD, we propose a stage-wise distillation strategy in our PCD framework. Different from existing methods which perform all-class distillation, we develop a bidirectional stage-wise distillation mechanism with a progressive Fine-to-Coarse Learning (F2CL) for knowledge accumulation and reverse Coarse-to-Fine Learning (C2FL) for knowledge refinement. 

During the bidirectional learning processes, we divide the sequence-guided LD into multiple stages and class-level distillation are separated into different groups in each stage. For notation, we introduce a hyperparameter $S$ to denote the number of stages and class groups in each distillation stage as:  
%Specifically, we introduce a hyperparameter \(S\) to control the size of class subset used in each step of distillation. For the fine-to-coarse process and coarse-to-fine process, size of class subset size at each step can be represented as:
\begin{equation}
    m^{F2CL}_i\ = \frac{C}{S - i + 1}, \quad m^{C2FL}_i\ = \frac{C}{i}, \quad i=1,2,...,S
\end{equation}
where $m_i$ denotes the size of class groups in $i^{th}$ stage. With $S=3$, for instance, group sizes of respective stages are $\frac{C}{3}$, $\frac{C}{2}$, and $\frac{C}{1}$ during the F2CL process. In other words, class-level distillation are separately conducted in three class subsets in the first distillation stage, which progressively propagates to the last stage in which class-level ensemble distillation is performed.
%Taking \(S\) = 3 as an example, the logits are partitioned into groups of different sizes at each step. In the fine-to-coarse process, the group size of each stage is $\frac{C}{3}$, $\frac{C}{2}$, and $\frac{C}{1}$, respectively. In the coarse-to-fine process, the group sizes are reversed accordingly. We use Eq.~\eqref{Eq5} to define the class index set of the \( j \)-th group as \( B^{f\rightarrow c}_{i,j} = \{ \mathbf{I}_k \mid k \in [(j-1) \cdot m^{f\rightarrow c}, j \cdot m^{f\rightarrow c}] \} \) for fine-to-coarse process and \( B^{c\rightarrow f}_{i,j} = \{ \mathbf{I}_k \mid k \in [(j-1) \cdot m^{c\rightarrow f}, j \cdot m^{c\rightarrow f}] \} \) for coarse-to-fine process.

%Next, we will explain from the fine-to-coarse process. The coarse-to-fine process can be obtained by simply replacing \(B^{f\rightarrow c}_{i,j}\) with \(B^{c\rightarrow f}_{i,j}\).
In each distillation stage, teacher-student logit alignment in different groups is achieved in sequence from challenging class groups to easier ones. In particular, when performing class-level distillation in $j^{th}$ group of $i^{th}$ stage, logits of other groups are not considered by the following logit masking strategy:

%Then we mask the student logits \( z^s \) and teacher logits \( z^t \), keeping only the logits within the corresponding group (others are set to \(-\infty\)):

\begin{equation}
    z^{s}_{i,j} = \text{mask}(z^s, I_{i,j}), \quad z^{t}_{i,j} = \text{mask}(z^t, I_{i,j})
\end{equation}
where \(\text{mask}(\cdot)\) denotes the operation of masking out the classes beyond the index set $I_{i,j}$, assigning their logits with $-\infty$ such that they are discarded in subsequent processing. Then, the softmax calculation is applied to masked logits, yielding probability distributions of $j^{th}$ class group at $i^{th}$ stage:
\begin{equation}
    p_{i,j} = \text{softmax}\left(\frac{z^{s}_{i,j}}{\tau}\right), \quad
    q_{i,j} = \text{softmax}\left(\frac{z^{t}_{i,j}}{\tau}\right)
\end{equation}

In addition, an adaptive weighting mechanism is introduced into group-wise logit alignment, allowing the student to focus on the classes that are inadequately learned for further improving distillation performance:
\begin{equation}
    \lambda_{i,j} = 1 -  \cos \left( p_{i,j}, q_{i,j} \right)
\end{equation}
where $\lambda$ is essentially cosine distance metric that measures the teacher-student discrepancy. Thus, weighted distillation loss for $j^{th}$ group of $i^{th}$ stage is calculated as:
\begin{equation}\label{Eq:aw}
    \mathcal{D}_{i,j} =\lambda_{i,j} \cdot \mathrm{KL}(p_{i,j} \parallel q_{i,j})\cdot \tau^2
\end{equation}

The complete distillation loss for progressive fine-to-coarse logit transfer is formulated as:
\begin{equation}
    \mathcal{L}_{KL} = \sum_{i=1}^{S} \sum_{j=1}^{M}  \mathcal{D}_{i,j}
\end{equation}
where $M$ denotes the number of class subsets in each stage.

To facilitate knowledge refinement, reverse coarse-to-fine distillation is also performed similar to F2CL procedure, such that bidirectional logit alignment is achieved for comprehensive knowledge transfer. The overall loss function can be expressed as:
\begin{equation}
    \mathcal{L} = \mathcal{L}_{\mathrm{CE}} + \alpha \cdot (\mathcal{L}^{F2CL}_{KL} + \mathcal{L}^{C2FL}_{KL})
\end{equation}
where \(\alpha\) is a weighting parameter that balances the cross-entropy loss and LD loss.

\section{Experiments}\label{sec4}
\subsection{Experimental Setup}
\subsubsection{Datasets and evaluation metrics.} 
To evaluate our proposed method, we have conducted extensive experiments on public benchmarking datasets including CIFAR~\cite{krizhevsky2009learning}, ImageNet~\cite{russakovsky2015imagenet} and MS-COCO~\cite{lin2014microsoft} for both classification and detection tasks. In terms of performance measure, we report Top-1(\%) and Top-5(\%) accuracies for classification along with average precision (AP) for detection.
%Among them, CIFAR-100 and ImageNet are used for classification while MS-COCO for detection task.

%\subsubsection{Implementation details.} DCDS has two hyper-parameters, \(levels\) and balance parameter \(\alpha\). The detailed analysis of \(levels\) can be seen in the ablation study. Specifically, \(levels\) is set as 3 for the distillation tasks with different architectures, and 5 for the distillation tasks with similar architectures. As for \(\alpha\), for a fair comparison with previous logit-based methods, we follow the same setting used in the original KD and DKD methods.

\subsubsection{Implementation details.} 
In our experiments, different classic deep convolutional networks are involved, e.g., VGG~\cite{simonyan2014very}, ResNet~\cite{he2016deep}, MobileNet~\cite{howard2017mobilenets,sandler2018mobilenetv2}. Following DKD in implementation, the batch size is set as 64, while initial learning rate is 0.01 for ShuffleNet~\cite{ma2018shufflenet} and MobileNet-V2, and 0.05 for the other networks (including VGG, WRN~\cite{zagoruyko2016wide} and ResNet) in CIFAR dataset. Besides, the number of training epochs is set to 240, the learning rate is divided by 10 at 150, 180, and 210 epoch with a 20-epoch warm-up period to accelerate convergence. In ImageNet, models are trained with 100 epochs. The batch size is set to 512, whilst the learning rate starts at 0.2 and is divided by 10 every 30 epochs. In MS-COCO, the training epoch is 12. Regarding the setting of two hyper-parameters $S$ and balancing parameter $\alpha$ in our framework, they will be discussed in the following parameter analysis. In our comparative studies, competing methods incorporate state-of-the-art FD~\cite{romero2015fitnetshintsdeepnets,park2019relational,tian2019contrastive,heo2019comprehensive,verma2022re,zagoruyko2016paying} and LD~\cite{hinton2015distillingknowledgeneuralnetwork,li2023curriculum,zhao2022decoupled,wen2021preparing,cao2023excellent,lan2025improve} methods.  %The classification task based on ImageNet are conducted in two RTX 4090 GPUs, while other experiments are performed in a RTX 3090 GPU.

\begin{table}[!ht]
    \centering
    \caption{Top-1(\%) results on CIFAR validation when teachers and students have the \textbf{same} network architectures. The best and suboptimal results of LD are highlighted in \textbf{bold} and \underline{underlined}, respectively. For the cases when the best FD result surpasses the best LD counterpart, the corresponding results are highlighted with *. \(\Delta\) indicates the performance improvement of our method over the vanilla KD. All the results are obtained by averaging three trials.}
    \begin{tabular}{cccccccc}
    \hline
        \multirow{4}{*}{\begin{tabular}[c]{@{}c@{}}Type \end{tabular}}
        ~ & Teacher & ResNet32×4 & VGG13 & WRN-40-2 & ResNet56 & ResNet110 & ResNet110  \\ 
        ~ & ~ & 79.42 & 74.64 & 75.61 & 72.34 & 74.31 & 74.31  \\ 
        ~ & Student & ResNet8×4 & VGG8 & WRN-16-2 & ResNet20 & ResNet32 & ResNet20  \\ 
        ~ & ~ & 72.50  & 70.36 & 73.26 & 69.06 & 71.14 & 69.06  \\  \hline
        \multirow{5}{*}{\begin{tabular}[c]{@{}c@{}}FD \end{tabular}}
        ~ & FitNet~\cite{romero2015fitnetshintsdeepnets} & 73.50  & 71.02 & 73.58 & 69.21 & 71.06 & 68.99  \\ 
        ~ & RKD~\cite{park2019relational} & 71.90  & 71.48 & 73.35 & 69.61 & 71.82 & 69.25  \\ 
        ~ & CRD~\cite{tian2019contrastive} & 75.51 & 73.94 & 75.48 & 71.16 & 73.48 & 71.46  \\ 
        ~ & OFD~\cite{heo2019comprehensive} & 74.95 & 73.95 & 75.24 & 70.98 & 73.23 & 71.29  \\ 
        ~ & ReviewKD~\cite{verma2022re} & 75.63 & 74.84* & 76.12 & 71.89* & 73.89* & 71.34  \\ \hline
        \multirow{5}{*}{\begin{tabular}[c]{@{}c@{}}LD \end{tabular}}
        ~ & KD~\cite{hinton2015distillingknowledgeneuralnetwork} & 73.33 & 72.98 & 74.92 & 70.66 & 73.08 & 70.67  \\ 
        ~ & CTKD~\cite{li2023curriculum} & 73.39 & 73.52 & 75.45 & 71.19 & 73.52 & 70.99  \\ 
        ~ & DKD~\cite{zhao2022decoupled} & \underline{76.19} & \textbf{74.68} & \underline{75.70}  & 71.46 & \underline{73.76} & 71.28  \\ 
        ~ & LA~\cite{wen2021preparing} & 73.46 & 73.51 & 74.98 & 71.24 & 73.39 & 70.86  \\ 
        ~ & RC~\cite{cao2023excellent} & 74.68 & 73.37 & 75.43 & \underline{71.59} & 73.44 & \underline{71.41} \\
        ~ & LR~\cite{lan2025improve} & 76.06 & 74.36 & 75.62 & 70.74 & 73.52 & 70.61 \\
        ~ & \textbf{PCD(Ours)} & \textbf{76.34} & \underline{74.40}  & \textbf{76.15} & \textbf{71.63} & \textbf{73.83} & \textbf{71.89}  \\ \hline
         ~ & $\Delta$ & \(+3.01\) & \(+1.42\) & \(+1.23\) & \(+0.97\) & \(+0.75\) & \(+1.22\) \\ \hline
    \end{tabular}
    \label{table1}
\end{table}

\subsection{Main Results}
\begin{table}[!ht]
    \centering
    \caption{Top-1(\%) results on CIFAR validation when teachers and students have \textbf{different} network architectures. The notations are consistent with those in Table~\ref{table1}.}
    \begin{tabular}{cccccccc}
    \hline
        \multirow{4}{*}{\begin{tabular}[c]{@{}c@{}}Type \end{tabular}}
        ~ & Teacher & ResNet32×4 & ResNet32×4 & WRN-40-2 & WRN-40-2 & VGG13 & ResNet50  \\ 
        ~ & ~ & 79.42 & 79.42 & 75.61 & 75.61 & 74.64 & 79.34  \\ 
        ~ & Student & SHN-V2 & WRN-16-2 & ResNet8×4 & MN-V2 & MN-V2 & MN-V2  \\ 
        ~ & ~ & 71.82  & 73.26 & 72.50  & 64.6 & 64.6 & 64.60   \\  \hline
        \multirow{5}{*}{\begin{tabular}[c]{@{}c@{}}Feature \end{tabular}}
        ~ & FitNet & 73.54  & 74.70  & 74.61 & 68.64 & 64.16 & 63.16  \\ 
        ~ & RKD & 73.21  & 74.86 & 75.26 & 69.27 & 64.52 & 64.43  \\ 
        ~ & CRD & 75.65 & 75.65 & 75.24 & 70.28 & 69.73 & 69.11  \\ 
        ~ & OFD & 76.82 & 76.17* & 74.36 & 69.92 & 69.48 & 69.04  \\ 
        ~ & ReviewKD & 77.78* & 76.11 & 74.34 & 71.28* & 70.37* & 69.89  \\ \hline
        \multirow{5}{*}{\begin{tabular}[c]{@{}c@{}}Logit \end{tabular}}
        ~ & KD & 74.45 & 74.90  & 73.97 & 68.36 & 67.37 & 67.35  \\ 
        ~ & CTKD & 75.37 & 74.57 & 74.61 & 68.34 & 68.50  & 68.67  \\ 
        ~ & DKD & \underline{76.56} & 75.70 & \underline{75.56} & 68.88 & 69.45 & 70.27  \\ 
        ~ & LA & 75.14 & 74.68 & 73.88 & 68.57 & 68.09 & 68.85  \\ 
        ~ & RC & 75.61 & 75.17 & 75.22 & 68.72 & 68.66 & 68.98 \\
        ~ & LR & 76.27 & \underline{76.10} & 75.26 & \underline{69.02} & \underline{69.51} & \underline{70.31} \\
        ~ & \textbf{PCD(Ours)} & \textbf{77.46} & \textbf{76.13} & \textbf{76.81} & \textbf{69.63} & \textbf{69.57} & \textbf{70.38}  \\ \hline
        ~ & $\Delta$ & \(+3.01\) & \(+1.23\) & \(+2.84\) & \(+1.27\) & \(+2.20\) & \(+3.03\) \\ \hline
    \end{tabular}
    \label{table2}
\end{table}

\begin{table}[!ht]
    \centering
    \caption{Results on the ImageNet validation. In implementation, ResNet-50 and MobileNet-V1 are respectively utilized as the teacher and the student model. All the results are obtained by averaging three trials.}
    \begin{tabular}{ccc|cccc|cccc}
    \hline
        ~ & Type & ~ & \multicolumn{4}{c|}{FD} & \multicolumn{4}{c}{LD} \\ \hline
        ~ & Teacher & Student & AT & OFD & CRD & ReviewKD & KD & RC & LR & \textbf{PCD (Ours)} \\ 
        Top-1(\%) & 76.16 & 68.87 & 69.56 & 71.25 & 71.37 & 72.56* & 70.50 & \underline{71.86} & 71.76 & \textbf{71.98}  \\ 
        Top-5(\%) & 92.86 & 88.76 & 89.33 & 90.34 & 90.41 & 91.00*  & 89.80 & 90.54 & \textbf{90.93} & \underline{90.74}  \\ \hline
    \end{tabular}
    \label{table3}
\end{table}

\begin{table}[!ht]
    \centering
    \setlength{\tabcolsep}{8pt} % 设置列间距
    \renewcommand{\arraystretch}{1.1} % Adjust row height
    \caption{Results on MS-COCO with Faster-RCNN used for the detector. Different detector backbones constitute pairwise teacher-student models, namely ResNet-101 \& ResNet-18 and ResNet-50 \& MobileNet-V2.}
    \begin{tabular}{c|ccc|ccc}
    \hline
        ~ & \multicolumn{3}{c|}{ResNet-101 \& ResNet-18} & \multicolumn{3}{c}{ResNet-50 \& MobileNet-V2} \\ \hline
        ~ & $AP$ & $AP_{50}$ & $AP_{75}$ & $AP$ & $AP_{50}$ & $AP_{75}$  \\
        Teacher & 42.04 & 62.48 & 45.88 & 40.22 & 61.02 & 43.81  \\ 
        Student & 33.26 & 53.61 & 35.26 & 29.47 & 48.87 & 30.90   \\  \hline
        KD & 33.85 & 54.53 & 36.58 & 30.08 & 50.28 & 31.35  \\ 
        FitNet & 34.08 & 54.11 & 36.48 & 30.15  & 49.80  & 31.69  \\ 
        \textbf{PCD (Ours)} & \textbf{34.16} & \textbf{55.98} & \textbf{36.60}  & \textbf{30.22} & \textbf{52.44} & \textbf{31.75}  \\ \hline
    \end{tabular}
    \label{table4}
\end{table}

\subsubsection{Image classification on CIFAR.}
%The top-1 validation accuracy(\%) comparison results of DCDS and other distillation approaches are reported in Table~\ref{table1}(homogeneous distillation pairs) and Table~\ref{table2}(heterogeneous distillation pairs). We can see that DCDS is either the optimal or suboptimal logit distillation method in all cases, and even better than feature-based methods in some cases. We note that compared with homogeneous distillation, heterogeneous distillation shows a more significant improvement. Since it can alleviate the difficulty of knowledge absorption caused by the capacity gap between the teacher and student with different architectures.
Table~\ref{table1} compares our method and other distillation approaches with varying settings of homogeneous teacher-student networks using Top-1(\%) accuracy. It can be observed that our method can significantly boost the vanilla LD (baseline) with dramatic performance gains. For example, the proposed PCD reports 76.34\% accuracy which significantly improves the baseline by 3.01\% when ResNet32$\times$4 and ResNet8$\times$4 are respectively used as the teacher and the student. Furthermore, compared to the other LD methods, PCD reports the best results except for the slight inferiority to DKD when VGG13 and VGG8 are adopted. The superiority of PCD is more impressive when heterogeneous models are involved, demonstrating that it consistently beats the other competitors and enormously improves the baseline as shown in Table~\ref{table2}.

\subsubsection{Image classification on ImageNet.}
%Top-1 and Top-5 accuracies of image classification on ImageNet is reported in Table~\ref{table3}. Compared with traditional knowledge distillation methods, our approach improves top-1 accuracy by 3.4(\%) and top-5 accuracy by 1.76(\%). However, it is slightly inferior to the DKD method. The reason for this discrepancy, we believe, is that ImageNet contains many similar classes, which results in very small probability differences among these similar categories in the logits, thereby affecting the reliability of our sorting~\cite{b34}.
In addition to the evaluations on CIFAR, we have also assessed our framework on larger ImageNet dataset. Consistent with the results on CIFAR, our method dramatically improves the conventional LD by approximately 1.5\% Top-1 and 0.9\% Top-5 accuracies as shown in Table~\ref{table3}. Meanwhile, the advantages of our method against the other LD approaches can also be observed, suggesting that PCD consistently outperforms RC and LR by 0.12\% and 0.22\% in Top-1 accuracy. 

\subsubsection{Object detection on MS-COCO.}
%We extend the PCD method to the object detection task. As shown in Table~\ref{table4}, for R-101 \& R-18, our method outperforms the traditional LD method by 0.31\% in the AP metric compared to the KD method, and for R-50 \& MV2, our method outperforms by 0.14\%. Both of them are surpassing the traditional FD method. These results fully demonstrate that our method has strong transferability.

To further verify the effectiveness of the proposed method, we have evaluated our PCD on MS-COCO for downstream object detection task. Results are reported on val 2017. As shown in Table~\ref{table4}, our method exceeds conventional KD by 0.31\% and 0.14\% AP respectively with two different settings, which fully suggests the preferable adaptability of our approach.

\begin{table}[!ht]
    \centering
    \setlength{\tabcolsep}{6pt} % 设置列间距
    \renewcommand{\arraystretch}{1.1} % Adjust row height    
    \caption{Ablation studies on CIFAR (Top-1). Notably, our approach is reduced to vanilla KD when removing all the three components, leading to the results displayed in the first row. ``-'' indicates that merely using LDR module is equivalent to vanilla KD.}
    \begin{tabular}{ccccccc}
    \hline
        LDR  & F2CL & C2FL & 
        \makecell{ResNet110 \\ ResNet20} &
        \makecell{WRN-40-2 \\ WRN-16-2} &
        \makecell{ResNet32×4 \\ SHN-V2} &
        \makecell{WRN-40-2 \\ ResNet8$\times$4}  \\ \hline
        ~ & ~ & ~ & 70.67 & 74.92 & 74.45 & 73.97  \\ 
        \checkmark & ~ & ~ & - & - & - & -  \\ 
        ~ & \checkmark & ~ & 71.36 & 75.57 & 76.8 & 76.61  \\ 
        ~ & ~ & \checkmark & 70.93 & 75.57 & 76.49 & 76.52  \\ 
        \checkmark & \checkmark & ~ & 71.69 & 75.21 & 77.26 & 76.29  \\ 
        \checkmark & ~ & \checkmark & 71.77 & 75.7 & 76.34 & 76.54  \\ 
        ~ & \checkmark & \checkmark & 71.08 & 75.99 & 76.63 & 76.34  \\ 
        \checkmark & \checkmark & \checkmark & \textbf{71.89} & \textbf{76.15} & \textbf{77.46} & \textbf{76.81}  \\ \hline
    \end{tabular}
    \label{table5}
\end{table}

\begin{table}[!ht]
    \centering
    \setlength{\tabcolsep}{6pt} % 设置列间距
    \renewcommand{\arraystretch}{1.1} % Adjust row height        
    \caption{The impact of Weighted Distillation Mechanism (WDM) in our method (Top-1).}
    \begin{tabular}{ccccc}
    \hline
        \makecell{Teacher \\ Student} & 
        \makecell{ResNet110 \\ ResNet20} & 
        \makecell{WRN-40-2 \\ WRN-16-2} & 
        \makecell{ResNet32×4 \\ SHN-V2} & 
        \makecell{WRN-40-2 \\ ResNet8×4} \\ \hline
        w/o WDM & 71.46  & 75.92 & 76.86 & 76.25  \\ 
        w/ WDM & \textbf{71.89} & \textbf{76.15 }& \textbf{77.46} & \textbf{76.81}  \\ \hline
    \end{tabular}
    \label{table6}
\end{table}

\subsection{Ablation Study}
%For a better understanding of PCD, we conduct ablation studies on CIFAR-100. The heterogeneous teacher-student pair are ResNet110-ResNet20 and WRN-40-2-WRN-16-2 and homogeneous pair are ResNet32×4-SHN-V2 and WRN-40-2-ResNet8×4.
To delve into our PCD method, we have conducted extensive ablation studies on CIFAR to explore the effects of individual module on the model performance. More specifically, we investigate the contribution of each component in our method, including LDR, F2CL and C2FL. As shown in Table~\ref{table5}, the complete model incorporating all the three modules achieves the best performance in both homogeneous and heterogeneous distillation scenarios. When discarding LDR, significant performance decline of approximately 0.8\% can be observed when using pairwise models of ResNet101-ResNet20 and ResNet32$\times$4-ShuffleNetV2. This implies that difficult-to-easy logit alignment is conducive to progressive learning, thereby benefiting knowledge transfer. In addition, bidirectional distillation process can improve the model learning ability, achieving comprehensive knowledge transfer and further performance gains. In addition, we also explore the beneficial effects of weighted distillation mechanism (WDM) formulated in Eq.~(\ref{Eq:aw}). As shown in Table~\ref{table6}, WDM contributes to further performance improvements, which implies that it enables adaptive class-specific distillation for bridging the gap between the teacher and the student.

\begin{table}[!ht]
    \centering
    \setlength{\tabcolsep}{6pt} % 设置列间距
    \renewcommand{\arraystretch}{1.1} % Adjust row height     
    \caption{Model performance with varying values of S on CIFAR (Top-1).}
    \begin{tabular}{ccccc}
    \hline
        \makecell{Teacher \\ Student} & 
        \makecell{ResNet110 \\ ResNet20} & 
        \makecell{WRN-40-2 \\ WRN-16-2} & 
        \makecell{ResNet32×4 \\ SHN-V2} & 
        \makecell{WRN-40-2 \\ ResNet8×4} \\ \hline
        S=3 & \textbf{71.89}  & \textbf{76.15}  & 76.39 & 76.49  \\ 
        S=4 & 71.11 & 75.61 & 76.80  & 76.65  \\ 
        S=5 & 71.03 & 76.01 & \textbf{77.46} & \textbf{76.81}  \\ \hline
    \end{tabular}
    \label{table7}
\end{table}

\begin{table}[!ht]
    \centering
    \setlength{\tabcolsep}{6pt} % 设置列间距
    \renewcommand{\arraystretch}{1.1} % Adjust row height         
    \caption{Model performance with varying values of $\alpha$ on CIFAR for pairwise models of ResNet110 and ResNet20.}
    \begin{tabular}{c|cccccc}
    \hline
        $\alpha$ & 0.1 & 0.2 & 0.5 & 1.0  & 2.0  & 3.0   \\ 
        \hline
        Top-1(\%) & 71.68 & 71.24 & 71.22 & \textbf{71.89 }& 70.33 & 70.20   \\ 
        \hline
    \end{tabular}
    \label{table8}
\end{table}

%\subsubsection{Effect of sorting module, fine-to-coarse process and coarse-to-fine process.}
%We investigate the contributions of each component in our method, including the sorting module, fine-to-coarse process and coarse-to-fine process. As shown in Table~\ref{table4}, the fusion of the three achieves the best performance for both homogeneous and heterogeneous teacher-student pairs. This demonstrates the effectiveness of our method in grouping classes based on their difficulty for progressive distillation. Besides, sorting module, fine-to-coarse and coarse-to-fine consistently improve the performance of conventional KD, verifying their effectiveness.

\subsection{Parameter Analysis}
In this section, we will discuss the hyperparameters $S$ and $\alpha$ involved in our framework.

\subsubsection{Impact of S.}
%We set different number of levels to study the performance of logit distillation under different level settings. As shown in Table~\ref{table5}, teahcer-student pairs with homogeneous structures perform best when \(S\) = 3. In contrast, the teacher-student pairs with heterogeneous structures perform best when \(S\) = 5. This indicates that the teacher-student pairs with heterogeneous structures need more fine-grained division. This may be because the heterogeneous structure makes it difficult for student to imitate the teacher's knowledge. Besides, too many levels for homogeneous teacher-student pairs will bring redundant information since their logits are similar. Therefore, we set levels as 3 for the homogeneous distillation and and 5 for the heterogeneous distillation. 
In PCD, we have evaluated the model performance with varying numbers of distillation stages. As shown in Table~\ref{table7}, the highest Top-1(\%) results are obtained with $S=3$ when both the teacher and the student have the same architectures. In contrast, more distillation stages are required when heterogeneous network architectures are involved, indicating that $S=5$ leads to the best model performance. This also suggests that additional distillation costs are indispensable for adequate knowledge transfer due to larger discrepancy between heterogeneous models. Therefore, to avoid knowledge redundancy resulting from excessive stages and inadequate knowledge transfer with insufficient stages, $S$ is set as 3 and 5 for employing homogeneous and heterogeneous models respectively in our experiments, such that an accuracy-efficiency tradeoff can be achieved. 
%The results are shown in Table~\ref{table6}. We explore the impact of the Weighting Mechanism in our method, the experimental results show its effectiveness.

\subsubsection{Effect of $\alpha$.}
In addition to S, we have explored the impact of $\alpha$ on the model performance. As shown in Table~\ref{table8}, the best result is reported at 71.89\% Top-1 accuracy with $\alpha=1$ when ResNet110 and ResNet20 are teacher-student models, demonstrating that an optimal balance is achieved between the distillation loss and task-specific loss.

\begin{figure}[htbp]
    \centering
    \begin{subfigure}{0.49\textwidth}
        \includegraphics[width=\linewidth]{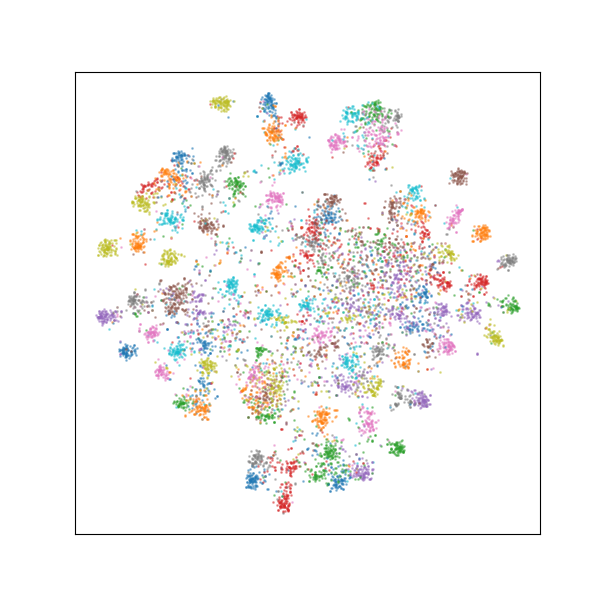}
%        \caption{KD}
        \label{fig:kd_tSNE}
    \end{subfigure}
    \hfill
    \begin{subfigure}{0.49\textwidth}
        \includegraphics[width=\linewidth]{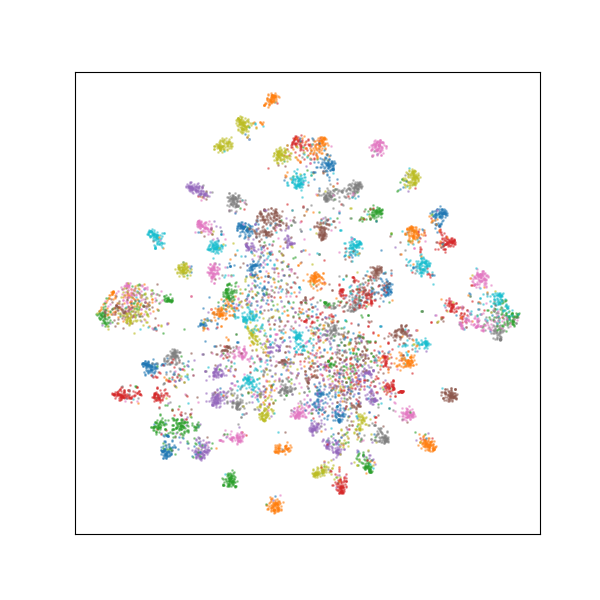}
%        \caption{PCD}
        \label{fig:bpbd_tSNE}
    \end{subfigure}
    \caption{t-SNE visualizations obtained by KD (left) and our PCD (right).}
    \label{fig3}
\end{figure}

\begin{figure}[htbp]
    \centering
    \begin{subfigure}{0.49\textwidth}
        \includegraphics[width=\linewidth]{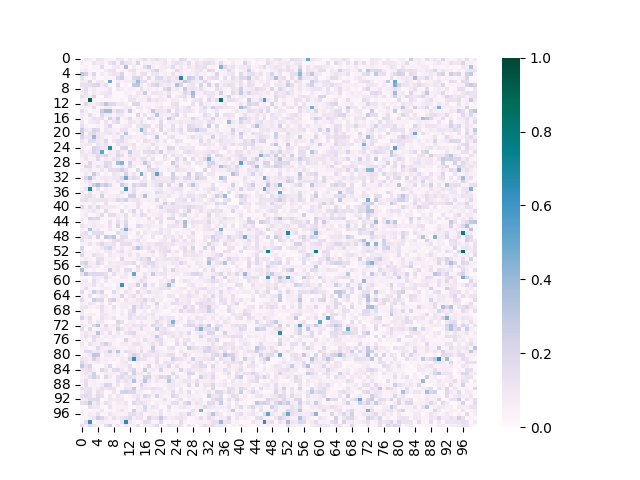}
        \label{fig:kd_tSNE}
    \end{subfigure}
    \hfill
    \begin{subfigure}{0.49\textwidth}
        \includegraphics[width=\linewidth]{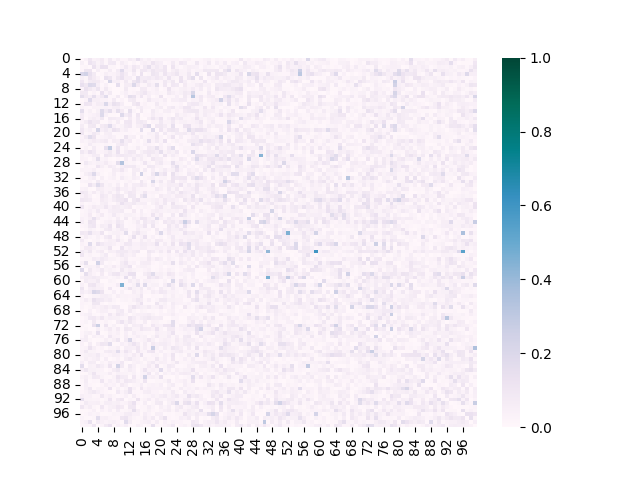}
        \label{fig:bpbd_tSNE}
    \end{subfigure}
    \caption{Visualized teacher-student logits difference in varying classes. Obviously, our PCD (right) leads to more consistent predictions than KD (left).}
    \label{fig4}
\end{figure}
%Difference of correlation matrices of student and teacher logits. Obviously, DCDS(right) leads to a smaller difference(more similar prediction) than KD(left).

\subsection{Visualizations}
%We present visualizations from two perspectives(with setting teacher as ResNet32×4 and student as ResNet8×4 on CIFAR-100). (1) The t-SNE(Fig.~\ref{fig2}) results show that representations of DCDS are more separable than KD, showing that the proposed DCDS can benefit students' discrimination ability. (2) We also visualize the difference of the correlation matrices of student and teacher logits (Fig.~\ref{fig3}). Compared with KD, DCDS helps the student to output more similar logits with the teacher, thus achieving better distillation performances.
In addition to the above quantitative results, we also intuitively compare the vanilla KD and our PCD via t-SNE~\cite{JMLR:v9:vandermaaten08a} visualizations and correlation matrices of teacher-student logit difference. As shown in Fig.~\ref{fig3}, our PCD allows more separable representations, which significantly improves the discriminative power of the student model. In addition, Fig.~\ref{fig4} illustrates that the proposed method leads to smaller prediction differences between the teacher and the student model, thereby effectively bridging their gap.

\section{Conclusion}\label{sec5}
%This paper reveals that treating logits as a whole may overlook those low-probability classes that still contain valuable information, thereby affecting the distillation effect. To overcome this limitation, we propose Difficult-guided Class-level Distillation with Sorted Logits(DCDS), which divides the logits into multiple class subsets for hierarchical distillation according to the difficulty of the categories. In addition, we divide the entire distillation process into two parts: fine-to-coarse and coarse-to-fine, allowing the student to follow a process of knowledge accumulation and verification, thereby fully learning and understanding the knowledge. Extensive experiments on several benchmark datasets demonstrate the effectiveness of DCDS both for image classification task and object detection task.

In this study, we reveal that low-confidence classes which contain discriminative information are likely to be downplayed in the traditional all-class distillation methods. To alleviate this drawback, we propose a progressive class-level distillation termed PCD to achieve step-by-step knowledge transfer. To be specific, our method includes two important modules, namely Logit Difference Ranking (LDR) and Bidirectional Stage-wise Distillation (BSD). The former generates class-level distillation sequence based on the teacher-student prediction difference such that the difficult classes are prioritized. Subsequently, BSD component achieves performs multi-stage knowledge transfer on the class sequence with progressive fine-to-coarse distillation for knowledge accumulation and reverse coarse-to-fine learning for knowledge refinement. Extensive experiments on both classification and detection tasks demonstrate the promise of our proposed method.

\bibliographystyle{splncs04}
\bibliography{reference}
\end{sloppypar}
\end{document}